\documentclass{article}
\usepackage{spconf,amsmath,graphicx}
\usepackage{hyperref}

\usepackage{xcolor}

\title{Towards Domain Adaptation from Limited Data \\ for Question Answering Using Deep Neural Networks}
\name{Timothy J. Hazen, Shehzaad Dhuliawala, Daniel Boies}
\address{Microsoft Research Montreal \\
         \tt{\{tj.hazen, shdhulia, daboies\}@microsoft.com}}

\begin{document}
%
\maketitle
\begin{abstract}
This paper explores domain adaptation for enabling question answering (QA) systems to answer questions posed against documents in new specialized domains. Current QA systems using deep neural network (DNN) technology have proven effective for answering general purpose factoid-style questions. However, current general purpose DNN models tend to be ineffective for use in new specialized domains. This paper explores the effectiveness of transfer learning techniques for this problem. In experiments on question answering in the  automobile manual domain we demonstrate that standard DNN transfer learning techniques work surprisingly well in adapting DNN models to a new domain using limited amounts of annotated training data in the new domain.
\end{abstract}

\section{Introduction}
\label{sec:intro}

\begin{figure}[th]
    \includegraphics[width=1.0
    \linewidth]{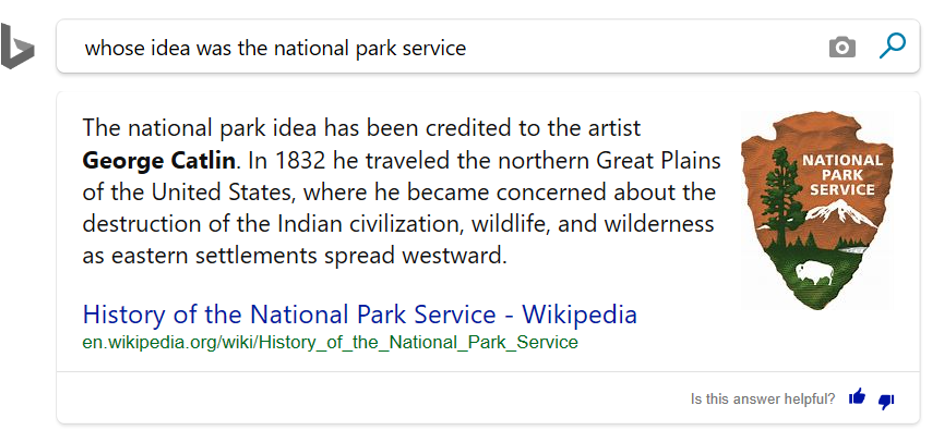}
    \caption{Example MRC-based answer produced by Bing for the question, "Whose idea was the National Park Service?". The answer is highlighted in bold text within the passage where it was found.}
    \label{fig:Bing-MRC-example}
\end{figure}

Conversational agents such as Siri and Alexa as well as traditional search engines such as Google and Bing have been steadily increasing the range and scope of user questions for which they can provide instant answers, i.e., direct answers to questions as opposed to links to web pages that may contain an answer. While early development of this capability focused on providing answers that could be extracted from structured databases or knowledge graphs, deep learning advances have enabled the capability to 
perform question answering (QA) using deep neural networks (DNNs) to extract an answer directly from a text passage. The use of DNNs for question answering is often alternatively referred to as machine reading comprehension (MRC). An example of an MRC-based answer produced by Bing is shown in Figure~\ref{fig:Bing-MRC-example}.

The release of the Stanford Question Answering Dataset (SQuAD)~\cite{rejpurkar-etal-2016-squad} in 2016 and the subsequent SQuAD 1.0 and SQuAD 2.0 competitions spurred impressive advances in MRC-based QA. Since the release of SQuAD, a variety of new modeling approaches including BiDAF~\cite{seo-bidaf-iclr-2017}, R-NET~\cite{wei-rnet-acl-2017}, QANet~\cite{yu-qanet-iclr-2018}, ELMo~\cite{peters-etal-2018-elmo}, BERT~\cite{devlin-etal-2019-bert}, RoBERTa~\cite{Roberta-Arxiv-2019}, MT-DNN\cite{MT-DNN-Arxiv-2019} and XLNET~\cite{yang_xlnet-2019-arxiv} have led to rapid improvements as reflected on the SQuAD leaderboard.\footnote{See: \url{https://rajpurkar.github.io/SQuAD-explorer}}

These recent research advances have also gained attention outside of the NLP research community. Press releases that touted the achievement of human parity on the SQuAD benchmark evaluations spurred industry interest in the development of QA systems for enterprise documents or document collections. Industry use cases include QA for corporate policies, technical manuals, legal documents and financial reports. While there has been extensive study in general purpose QA using MRC technology, the application of these techniques to specialized documents and use cases has received less attention. While it is hoped that a single general purpose model could perform QA robustly across many domains, our experiments have shown that today's state-of-the-art modeling techniques, trained with question and answer (Q\&A) pairs from multiple large general purpose data sets, are not yet robust and general enough to reliably answer questions in new specialized domains.

Since the release of the SQuAD dataset, there have been a variety of new data sets released covering different use cases and text data sources including NewsQA~\cite{trischler-etal-2017-newsqa}, SearchQA~\cite{Search-QA-Arxiv-2017}, TriviaQA~\cite{joshi-2017-triviaqa}, and MS-MARCO~\cite{MS-Marco-OpenReview-2018}. Most of these data sets have been fairly general in their topical content and are largely dominated by questions with short factoid-style answers (named entities, dates, objects, etc.). QA data sets in specialized domains are scarce with the most prominent ones focused on the medical domain, including emrQA~\cite{pampari-etal-2018-emrqa} and MEDIQA~\cite{ben-abacha-etal-2019-mediqa}.

In this paper, we present preliminary work towards the goal of performing QA against specialized documents. Our initial work in this area demonstrated that domain-specific MRC models can provide reliable and accurate question answering given enough in-domain data (e.g., 100K question and answer training examples). However, it is not always the case that a sufficiently large enough set of annotated Q\&A data will be available as such data is time-consuming and costly to collect. Thus, this paper considers adaptation to new specialized domains specifically for the scenario when the domain only has a limited number of annotated Q\&A pairs available for training purposes.  

To examine this problem area, our focus in this paper is specifically on QA for automobile manuals. We illustrate how this task is fundamentally different than the QA tasks of most general purpose data sets. We then demonstrate that standard transfer learning techniques work surprisingly well for domain adaptation from limited data in this scenario. We also present preliminary experiments that imply that applying unsupervised domain adaption techniques to a base model could provide some improvement in the absence of in-domain labeled training data, but that there may be no advantage to these methods once standard transfer learning methods are able to use even limited amounts of annotated training data in a new domain. 

\section{Related Work}

A wide variety of different techniques have been proposed for the general goal of domain adaptation, though only some have been tested within the context of MRC-QA systems. The most common approach for deep neural networks is to apply {\it transfer learning}, i.e. fine-tuning a pre-existing model using data in the new domain. The fine-tuning process can either use unsupervised or supervised training objectives depending on the availability of labeled data in the new domain.

For MRC-QA systems, transfer learning adapts the model to the new domain using the same standard error back-propagation training and supervised object function used to train the pre-existing model~\cite{Xu-BERT-Post-Training-2019}. To avoid over-fitting to the training data in the new domain, particularly when it is limited in size, the training is typically run with a small number of training epochs or with an appropriate early-stopping criteria. Augmenting the fine-tuning process with data similar to the target domain has also proven effective~\cite{Yang-BERT-Augmentation-2019}. 

Because the amount of available data for fine-tuning may be small and pre-trained models can be very large, there has been some concern about the efficiency of learning under these conditions. Using model compression techniques, it has been shown that the fine-tuning process can yield comparable performance improvement while training only a small fraction of the number parameters in the full model~\cite{houlsby-parameter-efficient-learning-2019-pmlr}. Though training efficiency is improved, this approach has not yet proven to help improve the accuracy of the resulting model in the new domain.

In cases where large amounts of unlabeled data in the new domain are available, fine-tuning a base model using a self-supervised language has been shown to improve modeling for a variety of domain specific tasks. Examples of this approach include ULMFiT~\cite{howard-ruder-2018-universal}, BioBERT~\cite{BioBERT-2019}, and SciBERT~\cite{SciBERT-2019}.  As far as we know, these techniques have not yet been shown to be effective for MRC-QA tasks.

Other techniques strive to make general models more robust to new domains or tasks. Feature augmentation is a general technique for domain adaptation for any type of machine learning model~\cite{daume-domain-adaptation-2007-acl}. This general idea has proven effective in a neural network model when the augmenting features are domain tags~\cite{yang-etal-2017-domain-adaptive-nets}. Importance weighting can also be used to reweight the pre-existing source training material to attempt to better match the characteristics of the target domain~\cite{plank-etal-2014-importance,Xia-ijcai-2018}. Improving domain robustness has also been explored using adversarial learning, which has proven successful for computer vision problems~\cite{Tzeng_2017_CVPR} but has thus far had less impact for question answering tasks~\cite{wang2019adversarial}.

\section{Datasets}

\begin{figure}[t]
    \centering
    \includegraphics[width=1.0\linewidth]{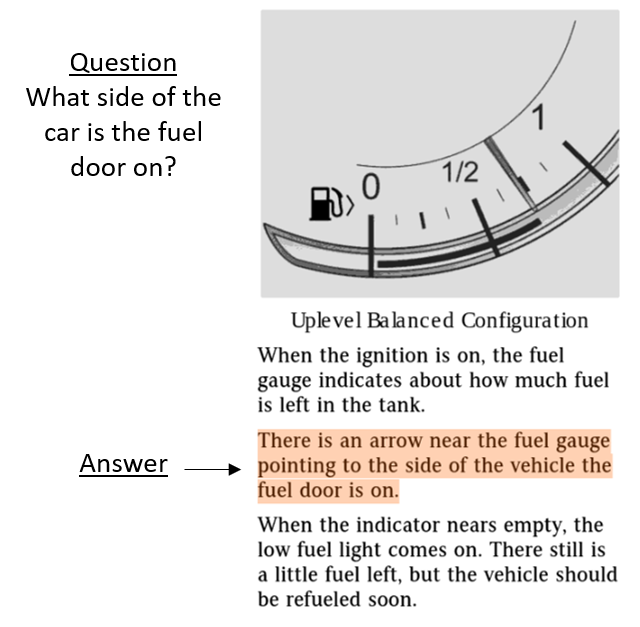}
    \caption{Example question and answer pair from the auto manual data set.}
    \label{fig:MRC-example}
\end{figure}

For our data we use question and answer (Q\&A) pairs where the answers are extracted from provided text passages. Our general purpose data sets are SQuAD, NewsQA and MS MARCO. For our specialized domain, we use data collected in the domain of automobile manuals. This data set contains questions collected against five different auto manuals. The text passages used for each question are the specific sections of the full manual that contain the most appropriate answer to each question. The reference answers are extracted portions of the section annotated as being the best answer to the question. Each of the manuals consists of several hundred sections and are typically several hundred pages in length.  An example Q\&A pair from a portion of one of the auto manual sections is shown in Figure~\ref{fig:MRC-example}.

For our domain adaptation experiments, we use data collected against a single BMW automobile manual. This data set consists 19K Q\&A pairs used for training and 1,937 held-out Q\&A pairs used for testing. Statistics of the sizes of the training and test sets for the other data sets used in our experiments are given in Table~\ref{tab:data_sets}. We use the publicly available development sets for SQuAD, NewsQA and MS MARCO as the test sets for experiments using these corpora. 


\begin{table}[t!]
\begin{center}
\begin{small}
\begin{tabular}{|l|c|c|} \hline
& \multicolumn{2}{c|}{Number of Q\&A Pairs}     \\ \cline{2-3}
Data Set        & Train Set & Test Set \\ \hline
SQuAD           & 88K       & 10,570       \\ \hline
NewsQA          & 93K       & 5,166        \\ \hline
MS MARCO        & 196K      & 15,365       \\ \hline
Auto (Full)     & 94K       & N/A          \\ \hline
Auto (BMW Only) & 19K       & 1,937        \\ \hline
\end{tabular}
\end{small}
\end{center}
\caption{MRC data sets used in this paper and their respective sizes. The test sets for SQuAD, NewsQA and MS MARCO in this paper are the publicly available development test sets. In the auto manual domain our experiments use a held test set of questions against a single BMW manual, though we use training data from four other automobile manuals in additional to the BMW manual data in some experiments. }\label{tab:data_sets}
\end{table}

\section{Preliminary Analyses}

\subsection{Evaluation Criteria}

This work focuses on the task of extractive QA, i.e., locating and extracting an appropriate text string from a long passage of text that directly answers a question. In our experiments, we use the MRC F1 measure defined for the SQuAD task as our evaluation metric~\cite{rejpurkar-etal-2016-squad}. This measure roughly measures the overlap of an answer against the correct answer using a measure akin to the F1 score used in information retrieval which combines measures of precision (i.e., the fraction of the proposed answer that is correct) and recall (i.e., the fraction of the correct answer that appears in the proposed answer).

Our evaluations assume that the correct document (or section from a document) containing the correct answer to the question has been been provided to the MRC component. In practice, a full end-to-end system will need to identify {\it both} the correct section of a document and the specific extracted answer within that section; but in this paper we focus on and evaluate only the second stage MRC component of the system. In our experiments we also only consider the case where every question has one and only one answer within a text passage. In practice, there may be instances where an appropriate answer to a question appears in multiple places in a document.

\subsection{Base QA Model}

The QA system in all experiments uses the pretrained BERT-base model feeding into a QA classification output layer.\footnote{Open source versions of BERT are available for PyTorch at \url{https://github.com/huggingface/transformers} and TensorFlow at \url{https://github.com/google-research/bert}} The BERT-base model is a transformer encoder that provides contextual word embeddings incorporating full self-attention across both the question and the full passage. It is pre-trained using self-supervision language modeling methods on a large corpus of unlabeled data available from the web including Wikipedia and a large corpus of books. The full QA model takes the concatenated question and text passage as the input, and its outputs the predicted starting and ending words of the answer from the passage. Model training using example Q\&A pairs both trains the output QA layers as well as fine-tuning the BERT-base model to the task.  

\subsection{Cross-Domain Performance Analysis}

Table~\ref{tab:cross_domain} shows MRC F1 performance when training and testing in a variety of within- and cross-domain scenarios. There are several interesting observations that can be made. First, when comparing performance of within-domain versus cross-domain models, performance not only drops in the cross-domain scenario but often drops by a large margin. This is particularly interesting in the cases of SQuAD, NewsQA and MS MARCO which are all topically diverse data sets. It is likely that differences in the style and structure of the text passages, the methodology under which the questions were collected, the types of questions being asked, and the labeling guidelines applied to the answers may all contribute to the cross-domain performance drop. 

We also examine the performance of a model trained using the union of the SQuAD, NewsQA, and MS MARCO dataset, which we refer to as our {\it general model}. It is interesting to note the performance of the general model on each of the individual domains has only small differences in performance relative to their respective domain specific models, resulting in a minor degradation on SQuAD and MS MARCO and a small improvement on NewsQA.  This implies the structural and stylistic differences between the domains is large enough that they blunt the value of the substantially larger number of Q\&A training pairs used to train the model.

The results for the domain specific models and the general model on the auto manual data are also interesting. Both the SQuAD model (F1=0.420) and NewsQA model (F1=0.422) perform relatively poorly on the auto manual data. By comparison the MS MARCO model (F1=0.681) performs significantly better and even outperforms the general model trained from all three general data sets (F1=.640). This implies the MS MARCO data is more similar to the auto manual scenario than SQuAD and NewsQA. All of the models learned from the general data perform significantly worse than the model learned solely from the full set of auto manual data. Learning a model from all four sets together (i.e. folding the auto manual training data in with the training data of the three general data sets) provides no difference in performance over the general model on the general data test sets, and provides only minimal improvements on the auto manual test set (F1=.852) versus just using the auto manual model (F1=.846). These results show domain differences between data sets can be large and learning general models from multiple domains may still not be sufficient for new specialized domains. 

\begin{table}[t!]
\begin{center}
\begin{small}
\begin{tabular}{|l|c|c|c|c|} \hline
& \multicolumn{4}{c|}{Test Set F1 Score}     \\ \cline{2-5}
Training Set & SQuAD & NewsQA & MARCO & Auto \\ \hline
SQuAD        & .880  & .502   & .453  & .420 \\ \hline
NewsQA       & .774  & .612   & .460  & .422 \\ \hline
MS MARCO     & .563  & .316   & .771  & .681 \\ \hline
Auto         & .223  & .169   & .562  & .846 \\ \hline
General      & .878  & .648   & .758  & .639 \\ \hline
All          & .878  & .648   & .758  & .852 \\ \hline
\end{tabular}
\end{small}
\end{center}
\caption{Experimental results for MRC-QA models trained and tested under a variety of within- and cross-domain scenarios.}\label{tab:cross_domain}
\end{table}

\subsection{Analysis of Domain Variation}

\begin{figure}[t]
    \centering
    \includegraphics[width=1.0\linewidth]{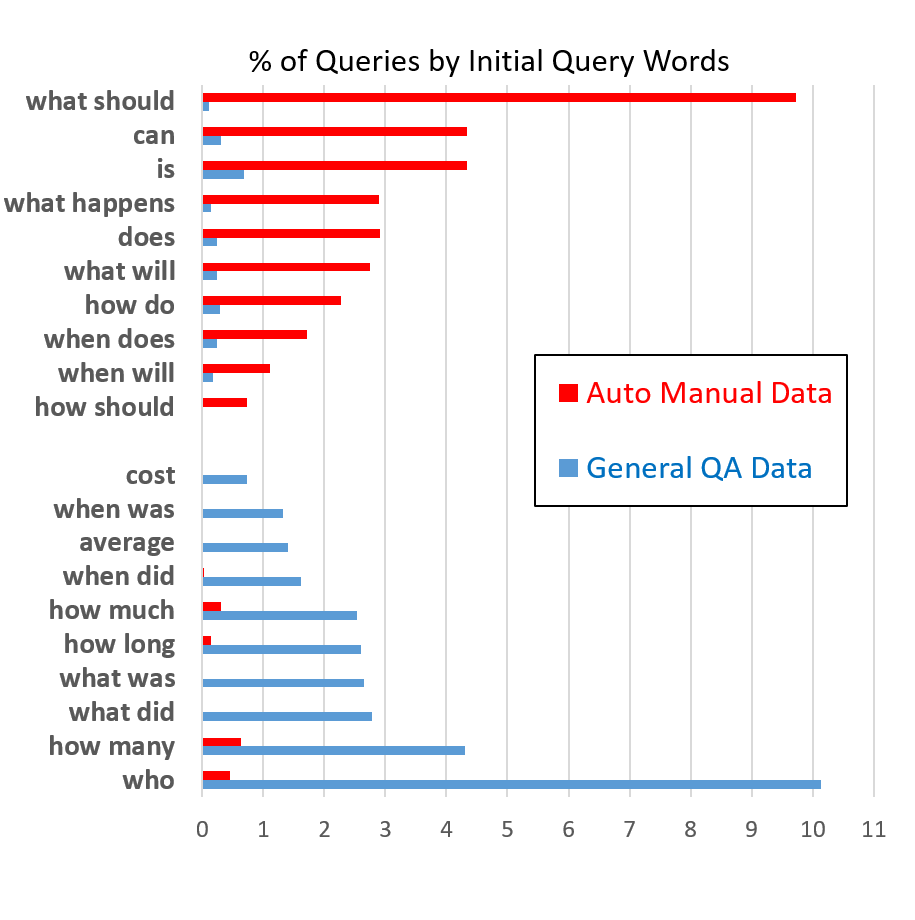}
    \caption{Common initial question phrases and their respective percentages with the general data sets (in blue) and the auto manual data sets (in red).}
    \label{fig:common_questions}
\end{figure}

The types of questions being asked in a domain have a direct impact on the types of answers that are required~\cite{Zhang-Question-Adaptation-2017}.  Clear differences can be observed between the most common question types contained in the general data (SQuAD, NewsQA and MS MARCO) from those in our auto manual data. Figure~\ref{fig:common_questions} highlights some of these differences. The left hand side of the figure shows common starting words or word bigrams contained in questions from the general data sets and the auto manual data set. The ten examples on the top of the table are common initial question words in the auto manual data that are less frequent in the general data. The bottom ten are initial question words that are common in the general data but much less frequent in the auto manual data. Common initial question words in the general data include {\it who}, {\it what was}, {\it how many}, {\it how long}, and {\it when did} that generally yield short factoid-style answers such as proper names, objects, numbers, or data/time expressions. These ten question types represent over 30\% of the questions in the general data but less than 3\% of the data in the auto manual data. On the other hand, the auto manual data has question types like {\it what should}, {\it what happens}, {\it how do} and {\it how should} that generally require longer answers such as instructions or technical descriptions. These 10 question types in the figure represent over 32\% of the questions in the auto manual data, but less than 2\% of the questions in the general data.

\begin{figure}[t]
    \centering
    \includegraphics[width=1.0\linewidth]{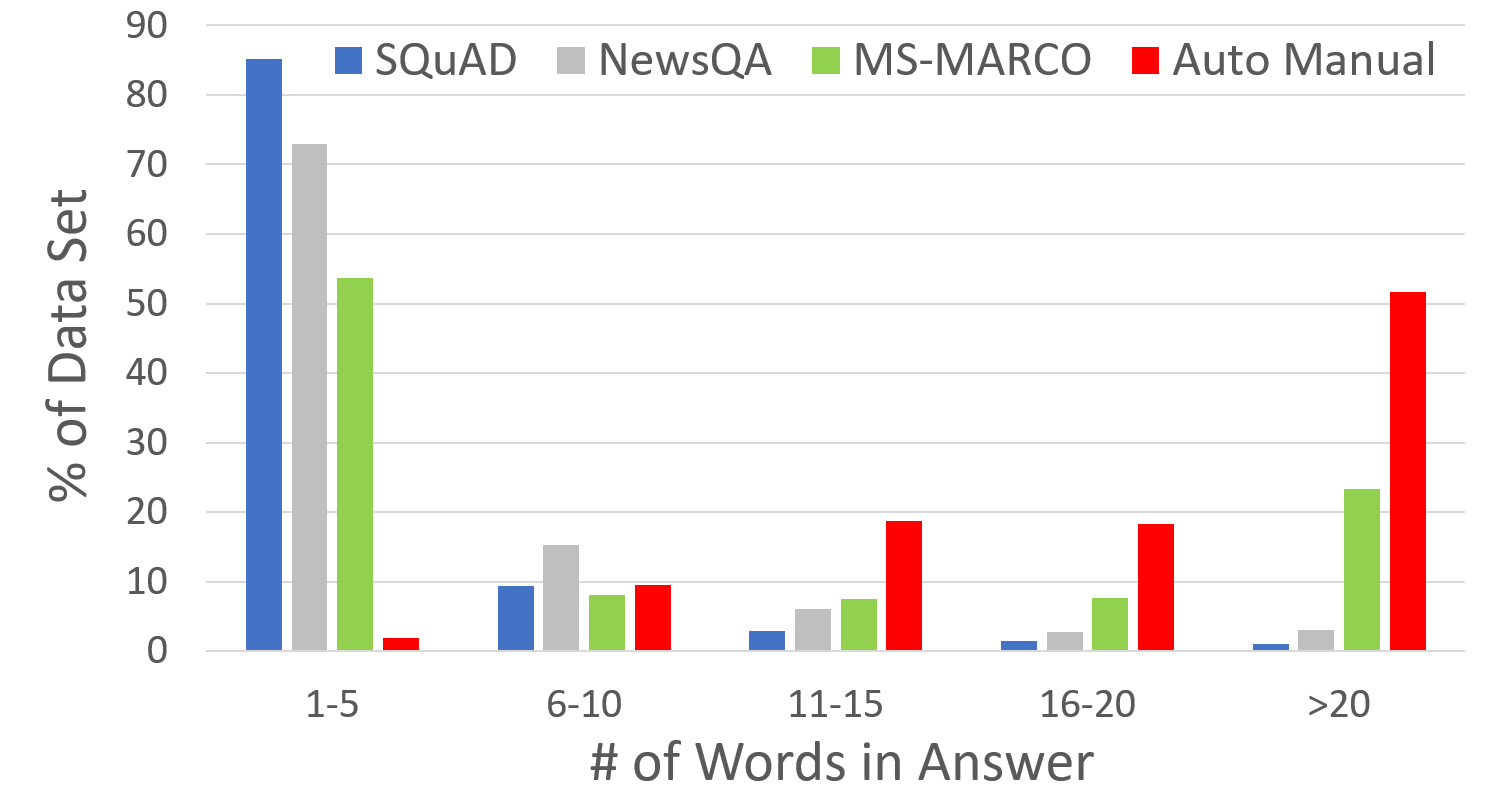}
    \caption{Answer length distributions for different Q\&A data sets over varying answer lengths.}
    \label{fig:answer_lengths}
\end{figure}

The differences in the question types between the general data and the auto manual data is also reflected in the length of the answers across these data sets. Figure~\ref{fig:answer_lengths} shows a histogram of answer lengths across the three corpora in the general data set and for the auto manuals. The SQuAD and NewsQA data sets are dominated by Q\&A pairs with short answers. Answers with only 5 words or less constitute 85\% of SQuAD Q\&A pairs and over 73\% of NewsQA. By contrast, over 50\% of the answers in the auto manual domain are greater than 20 words in length and less than 2\% are 5 words or less. MS MARCO has a more diverse set of answers; though a little more than 50\% of its answer are 5 words or less, more than 20\% of its answers are more than 20 words in length. The inclusion of a sizeable percentage of longer answers makes MS-MARCO more similar to the auto domain than SQuAD or NewsQA.

\section{Domain Adaptation Experiments}

\subsection{Experimental Conditions}

To evaluate domain adaptation from limited data, we have run a series of experiments adapting QA models trained on general data to the BMW auto manual data using only limited sets of randomly sampled questions from our training set. For adaptation, our BMW data set has approximately 19K total training samples, from which explore using smaller training set sizes of 1\%, 5\%, 10\%, 25\%, 50\% and 75\% of the full train set. A 1\% sample set corresponds to only 190 Q\&A pairs. We draw 5 different random sets for each of these sample sizes and report the average F1 score on our held out test set for each sample size.

\subsection{Standard Transfer Learning}

The domain adaptation in our initial experiments is performed with a standard transfer learning approach. For each sample set, we train a BERT-QA model using its standard back propagation training algorithm where our general QA model trained from SQuAD, NewsQA, and MS MARCO data serves as the starting point. Training is conducted for only 2 training epochs over each sampled set used for domain adaptation. 

\begin{figure}[t]
    \centering
    \includegraphics[width=1.0\linewidth]{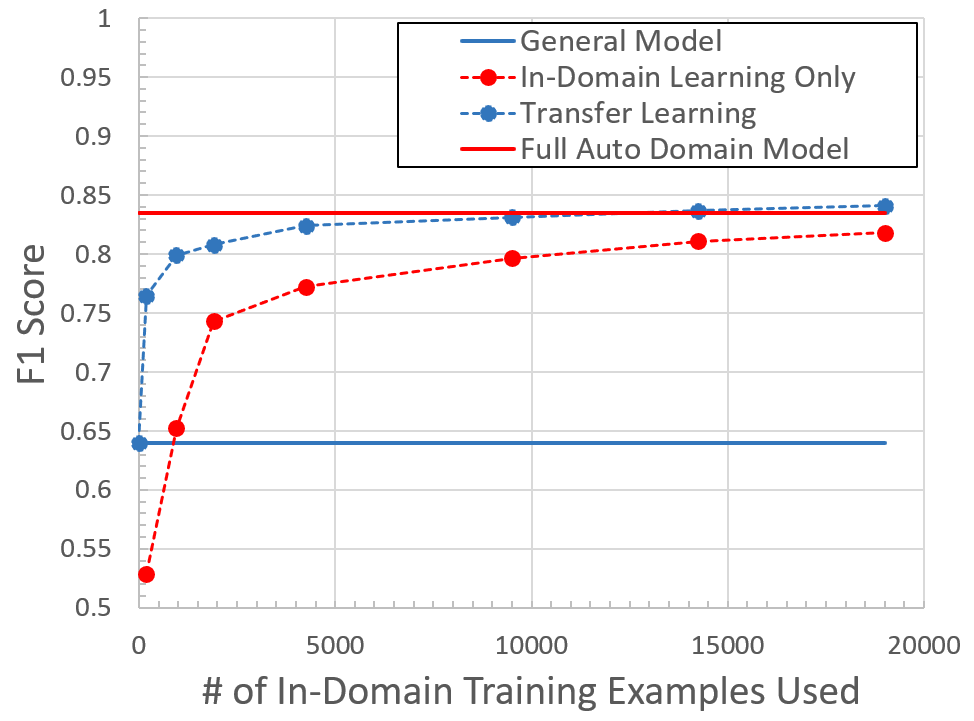}
    \caption{Q\&A F1 performance for variable amounts of in-domain BWM automobile manual training data when training with in-domain data only (in red) and with transfer learning of a general QA model (in blue).}
    \label{fig:transfer_learning}
\end{figure}

Our baseline domain adaptation experiments over variable sizes of available domain adaptation data are presented in Figure~\ref{fig:transfer_learning}.
In the figure, the F1 score for four different experimental conditions are shown as the amount of available adaptation/training data in auto manual domain is varied from 190 to 19K. The solid blue line with an F1 of 0.64 represents the accuracy of the general model. The solid red line represents a model trained from only the full set of 94K automobile manual Q\&A pairs.  The dotted red line represents training the QA model using variable amounts of BWM auto manual Q\&A pairs. The blue dotted line represents the application of transfer learning using the general QA model fine-tuned to variable amounts of the BMW auto domain data.

In Figure~\ref{fig:transfer_learning} we see that transfer learning is highly effective. The model's F1 performance is improved from 0.639 to over 0.765 using only 190 in-domain Q\&A pairs for  transfer learning. With only 950 Q\&A pairs for adaptation, the F1 improves to 0.799. Over the whole curve, we generally observe than the process of fine-tuning a general model using transfer learning requires approximately 10 times less data than a model trained only on in-domain data to achieve the same F1 score.

\begin{figure}[t]
    \centering
    \includegraphics[width=1.0\linewidth]{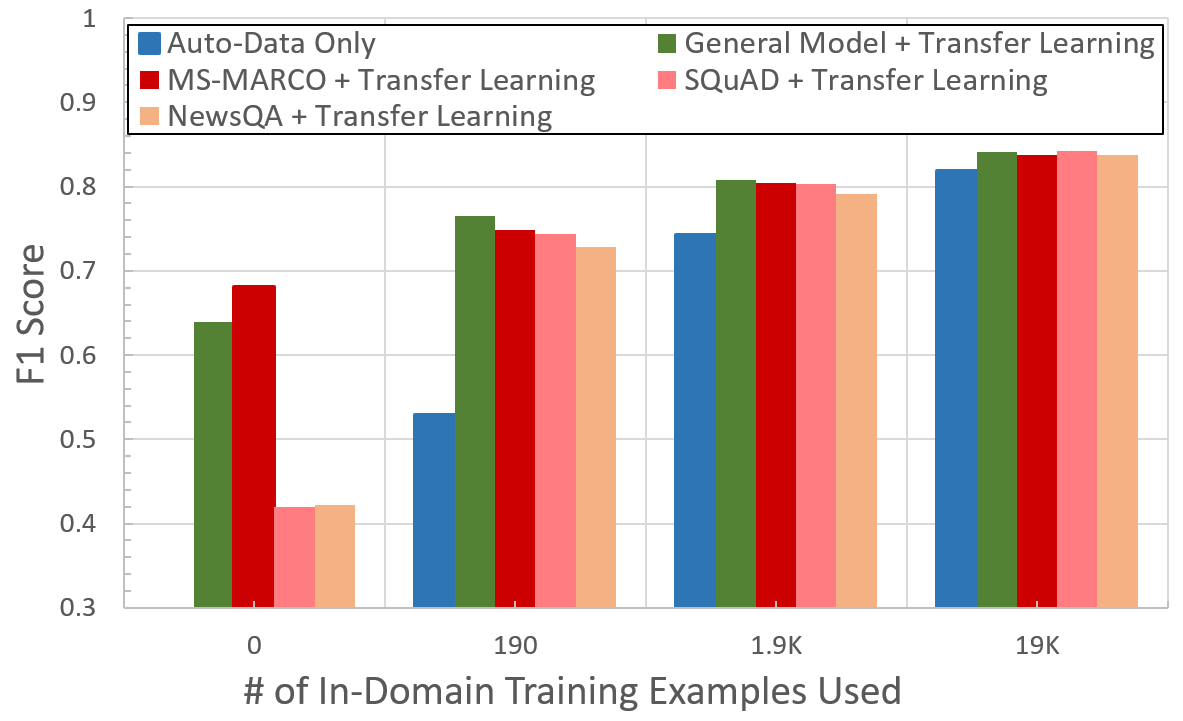}
    \caption{F1 results for transfer learning to the BWM automobile manual data using initial models trained from different QA data sets. }
    \label{fig:transfer_learning_bar}
\end{figure}

Recalling that the model trained on MS MARCO provides a better initial performance on the auto manual than the full general model, is is worth asking if that improved performance carries over when applying transfer learning. The bar chart in Figure~\ref{fig:transfer_learning_bar} shows that this is not the case. Despite the fact that the MS MARCO model performance (F1=0.681) is better out-of-the-box than the general model (F1=0.637), fine-tuning the general model with as little of 190 in-domain training examples yields better results (F1=0.765) than fine-tuning the MS MARCO model (F1=0.754). Also interesting is that fine-tuning the SQuAD and NewsQA models yields nearly the same results despite their poor initial performance out-of-the-box. 

The results in Figure~\ref{fig:transfer_learning_bar} appear to indicate that the fine-tuning process can overcome big domain mismatches even when using limited in-domain data for adaptation. The results further indicate that adding more Q\&A data into the base model training is better for the model during fine-tuning even when the added data is mismatched to the new domain. 

\subsection{Effect of Answer Length}

Given that there are large differences in the answer lengths between the general data sets used for training compared to the data in the auto manual domain, we sought to understand how much answer length itself affected both the base models and the ability of transfer learning to fine-tune to these differences. To explore this we made use of a technique called {\it importance weighting} in the training process of the general model~\cite{plank-etal-2014-importance,Xia-ijcai-2018}. Using this technique, different samples in the source training material are given different weights during the training of the base model
with the goal of making the {\it distribution} of data used during training resemble the distribution of data in the target domain.

To formalize this, importance weighting assigns a sample dependent weight to each data sample in the general data set. The weight is applied as a multiplicative factor on the learning rate used during the back-propagation training process.  We assume the best model learned from the source material is a model that tunes the weights on the source material in order to match the distributional characteristics of the target domain data as closely as possible. If we assume there is a feature $f$ (or possibly a multiple-dimension feature vector) that characterizes a training sample, we can achieve this distributional matching with this equation for the weighting function:
\begin{equation}
    w(f) = \frac{p_{t}(f)}{p_{s}(f)}
\end{equation}
Here $p_{t}(f)$ is the likelihood of observing a sample in the target domain with feature $f$, while $p_{s}(f)$ is the same likelihood function for the source domain (i.e., the general data set is the source domain in our case). When this weight is applied during training from Q\&A pairs from the source domain, it has the effect of enhancing samples from the source domain when the feature's value is more prominent in the target domain than the source domain, and suppressing samples where the feature's value is less prominent in the target domain. This technique can be applied to any feature that captures important distributional differences between the source and target domains. 

\begin{figure}[t]
    \centering
    \includegraphics[width=1.0\linewidth]{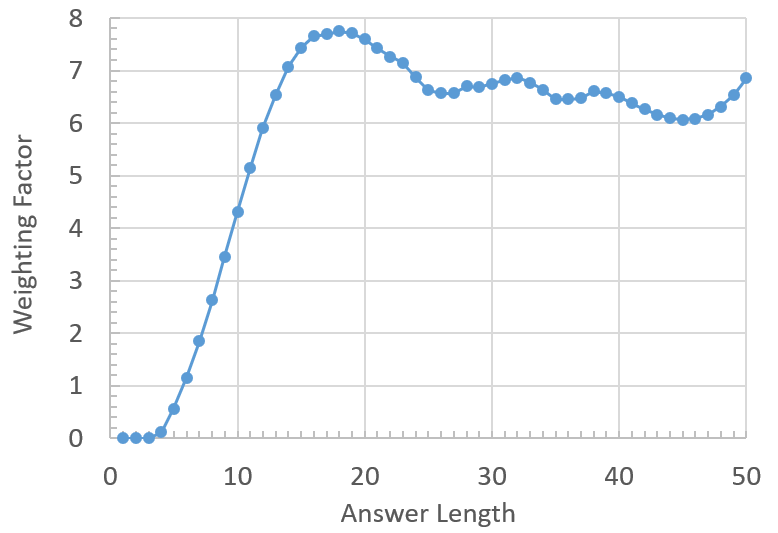}
    \caption{Example weighting function learned for the different Q\&A answer lengths when the answer length histogram is estimated from a random draw of 25\% of the target domain training data.}
    \label{fig:importance-weights}
\end{figure}

We explored the use of this technique with the answer length (in words) of a Q\&A pair as the feature $f$. We use a histogram estimation method to estimate the functions $p_{s}(f)$ and $p_{t}(f)$ where $f$ is the answer length.\footnote{For histogram estimation we use the histogram function in numpy using the {\it bins='auto'} setting.}  We need annotated data to do this estimation, so for the target domain of BWM we estimate the histograms using varying amounts of the target data to match the experimental paradigm used in Figure~\ref{fig:transfer_learning}. To avoid causing instability in the training process caused by weights that are excessively large due to major distributional differences between source and target, we cap the weight at a maximum value of 10. Figure~\ref{fig:importance-weights} shows an example weighting function learned from a random draw of 25\% of the target domain training data. Here we see Q\&A pairs in the source data with answer lengths of 6 words or longer are enhanced while Q\&A pairs with answers lengths of 5 words or less are suppressed during training. Training samples with answer lengths of greater than 10 words are enhanced by more than 5 times their original weight.

\begin{figure}[t]
    \centering
    \includegraphics[width=1.0\linewidth]{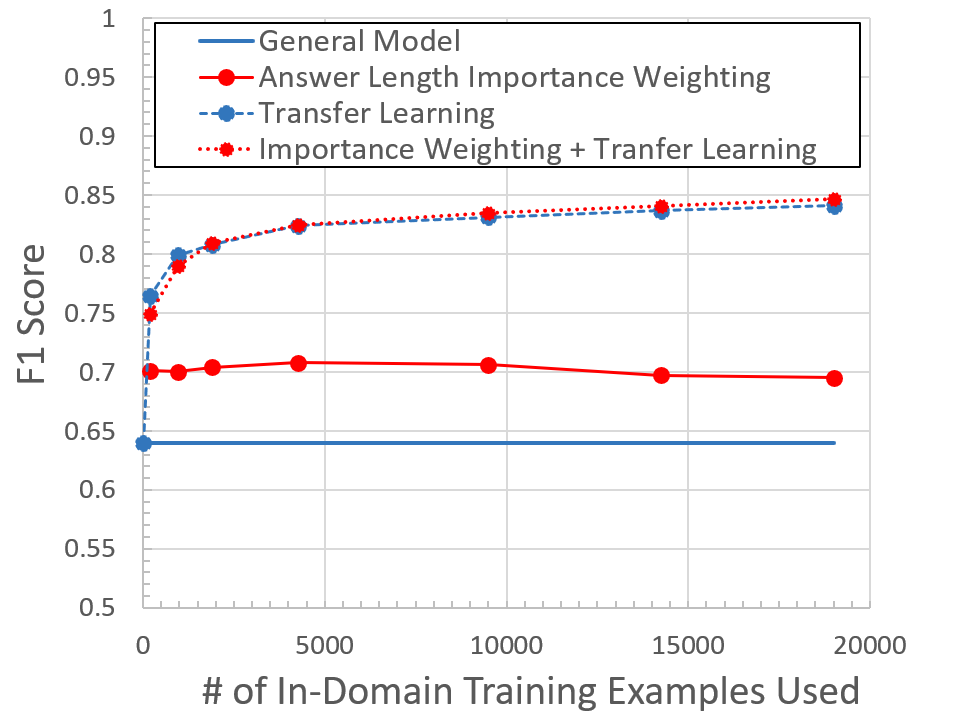}
    \caption{F1 results for transfer learning of an the importance weighted model compared to transfer learning of the baseline general model.}
    \label{fig:importance-weighting-f1}
\end{figure}

The solid red line in Figure~\ref{fig:importance-weighting-f1} shows the effect of applying importance weighting using varying amounts of in-domain training data to estimate $P_t(f)$. Note that no target data is used directly when training this model; only the weights on the original source data are changed to force the training to focus on source data whose answer lengths are more representative of the answer lengths in the target domain. The figure shows little difference in the resulting model based on the number of target examples used to estimate $P_{t}(f)$, with an F1 score of approximately 0.70 across the range. The score improvement from 0.64 to 0.70 using the importance weighting implies that answer length alone is an important factor in the difference between the general model and the domain specific model. However, adjusting the general model to specifically account for answer length differences only accounts for 30\% of the gap between the general model and the domain specific model. 

The dotted red line in Figure~\ref{fig:importance-weighting-f1} shows the results when the answer length adjusted general model is then fine-tuned to the in-domain data. In these experiments the fine-tuning data is always the same data that was used to estimate the length distribution for the importance weighting used to train the general model. It is interesting to note that the transfer learning curves for both the general model (the dotted blue line in Figure~\ref{fig:importance-weighting-f1}) and the answer length adjusted general model trained with importance weighting follow nearly identical tracks. Despite the improvement seen from the answer-length adjusted model over the original general model, the improvement disappears once transfer learning is applied, and in fact the answer-length adjusted model shows slightly worse performance than the original general model when fine-tuning on less than 1000 in-domain data samples. This mirrors the results in Figure~\ref{fig:transfer_learning_bar} where similar observations were made about the MS MARCO model. These results imply that the fine-tuning process of a BERT-QA general model using standard transfer learning techniques is able to efficiently overcome cross-domain differences in a model even with limited with limited in-domain data.

\section{Conclusion}
\label{sec:conclusion}

In examining the full set of results in this paper, these main conclusions can be drawn:
\begin{itemize}
    \item Deep learned question answering models trained from large amounts of general data may not perform adequately out-of-the-box on questions posed against documents in new specialized domains.
    \item QA models based on transformer encoders like BERT that are trained on large amounts of general data can be efficiently adapted to new domains with limited data.
    \item The efficiency of transfer learning of a general purpose QA model to a new domain is primarily affected by the amount of data used to train the general model and not on the initial accuracy of a pre-trained model. 
\end{itemize}
The surprising effectiveness of transfer learning of our BERT-based general QA model to the automobile domain using only limited amounts of annotated data the new domain gives us hope that a wide range of enterprise QA scenarios can be enabled without the requirement to collect large amounts of data in these domains. For future work we intend to verify these results on data for a range of new QA domains that we are currently in the process of collecting.



\end{document}